\documentclass[10pt,twocolumn, letterpaper]{article} 

\newcommand{\final}{1} 
\newcommand{\forReview}{0} 

\usepackage{cvpr}
\usepackage{times}
\usepackage{epsfig}
\usepackage{graphicx}
\usepackage{amsmath}
\usepackage{amssymb}
\usepackage{amsthm}
\usepackage{subcaption}
\usepackage{authblk}
\usepackage{cvpr}
\ifdefined\siggraph
\usepackage{times}
\fi

\usepackage{color}
\usepackage{ifthen}
\usepackage{float}
\usepackage{alltt}
\usepackage{newlfont} 
\usepackage{wrapfig}
\usepackage{booktabs}
\usepackage{multirow}

\usepackage{comment}
\usepackage{amsmath}
\usepackage{amssymb}
\usepackage{amsthm}
\usepackage{textcomp}

\definecolor{DeltaColor}{rgb}{0.039,0.73,0.71}
\definecolor{SetaColor}{rgb}{0.867, 0.0235, 0.376}
\definecolor{SigmaColor}{rgb}{0.98,0.45,0.0}
\definecolor{RedColor}{rgb}{0.8,0,0}
\definecolor{AlphaColor}{rgb}{0,0,0.8}
\definecolor{BetaColor}{rgb}{0.8,0,0.8}
\definecolor{GammaColor}{rgb}{0.5,0,0.7}
\definecolor{EpsilonColor}{rgb}{0.353,0.725,0.906}
\definecolor{TauColor}{rgb}{0.423,0.235,0.192}
\newcommand{\weikai}[1]{{\color{RedColor} Weikai: #1 $\qed$}}
\newcommand{\shichen}[1]{{\color{AlphaColor} Shichen: #1 $\qed$}}
\newcommand{\tianye}[1]{{\color{SigmaColor} Tianye: #1 $\qed$}}
\newcommand{\jun}[1]{{\color{GammaColor} Jun: #1 $\qed$}}
\newcommand{\yajie}[1]{{\color{DeltaColor} Yajie: #1 $\qed$}}
\newcommand{\hao}[1]{{\color{HaoColor} Hao: #1 $\qed$}}
\newcommand{\andrew}[1]{{\color{EpsilonColor} Andrew: #1 $\qed$}}
\newcommand{\paul}[1]{{\color{TauColor} Paul: #1 $\qed$}}

\newcommand{\warning}[1]{{\it\color{red} #1}}
\newcommand{\note}[1]{{\it\color{blue} #1}}
\newcommand{\nothing}[1]{}

\definecolor{AudioColor}{rgb}{0.56,0.34,0.62}

\definecolor{DeadlineColor}{rgb}{0.9,0.4,0} 
\newcommand{\deadline}[1]{{\bf\color{DeadlineColor} ETA: #1}}

\definecolor{figred}{rgb}{1,0,0}
\definecolor{figgreen}{rgb}{0,0.6,0}
\definecolor{figblue}{rgb}{0,0,1}
\definecolor{figpink}{rgb}{1,0.63,0.63}

\ifthenelse{\equal{\final}{1}}
{
\renewcommand{\shichen}[1]{}
\renewcommand{\weikai}[1]{}
\renewcommand{\tianye}[1]{}
\renewcommand{\jun}[1]{}
\renewcommand{\yajie}[1]{}
\renewcommand{\andrew}[1]{}
\renewcommand{\hao}[1]{}
\renewcommand{\paul}[1]{}
\renewcommand{\warning}[1]{}
\renewcommand{\note}[1]{}
\renewcommand{\deadline}[1]{}
}
{}

\newcounter{pccount}
\floatstyle{plain}

\newcommand{\filename}[1]{\url{#1}}
\newcommand{\foldername}[1]{\url{#1}}

\newcommand{\model}{soft rasterizer}

\newcommand{\modelshort}{SoftRas }

\newcommand{\aggregate}{\mathcal{A}}
\newcommand{\Dmap}{D}
\newcommand{\D}{D}
\newcommand{\pixel}{p}
\newcommand{\face}{f}
\newcommand{\dist}{d}
\newcommand{\height}{h}
\newcommand{\width}{w}
\newcommand{\sil}{\hat{S}}
\newcommand{\silGT}{S}
\newcommand{\silSmall}{\hat{S}}

\newcommand{\loss}{\mathcal{L}}
\newcommand{\mesh}{M}
\newcommand{\vertex}{v}
\newcommand{\inImg}{x}
\newcommand{\sig}{\delta}

\ifthenelse{\equal{\forReview}{1}}
{
	\usepackage[pagebackref=true,breaklinks=true,letterpaper=true,colorlinks,bookmarks=false]{hyperref}
	\def\cvprPaperID{****} 

	\ifcvprfinal\pagestyle{empty}\fi
}
{
	\usepackage[breaklinks=true,bookmarks=false]{hyperref}
	
	\cvprfinalcopy 
	
	\def\cvprPaperID{****} 

	\setcounter{page}{4321}
}

\def\cvprPaperID{4845} 

\ifcvprfinal\fi

\graphicspath{
	{figs/raster/}
	{figs/handdrawn/}
}

\begin{document}
	
\linespread{0.9}

\title{Soft Rasterizer: Differentiable Rendering 
for \\ Unsupervised Single-View Mesh Reconstruction}

\author[1,2]{Shichen Liu}
\author[1]{Weikai Chen}
\author[1,2]{Tianye Li}
\author[1,2,3]{Hao Li}
\affil[1]{USC Institute for Creative Technologies}
\affil[2]{University of Southern California}
\affil[3]{Pinscreen}
\affil[ ]{{\tt\small\{\href{mailto:lshichen@ict.usc.edu}{lshichen}, \href{mailto:wechen@ict.usc.edu}{wechen},	\href{mailto:tli@ict.usc.edu}{tli}\}@ict.usc.edu \quad \href{mailto:hao@hao-li.com}{hao@hao-li.com}}}

\twocolumn[{%
\renewcommand\twocolumn[1][]{#1}%
\maketitle
\vspace{-0.2in}
\nothing{
\begin{center}
    \centering
    \vspace{-15pt}
  \includegraphics[width=\linewidth]{teaserHolder.png}
    \captionof{figure}{Teaser.}
    \label{fig:teaser}
\vspace{-0.05in}
\end{center}%
}
}]


\nothing{
\note{
	Summary of Intro (make short and highlight the most important messages)
	\begin{itemize}
		\item What s the problem?
		Inverse graphics is key to reconstruct 3D shape from images. However, rendering is not differentiable due to the discrete rasterization operation. 
		\item What do we introduce?
		\item How does it roughly work? (main differentiator to existing methods? What's unique?)
		\item What can we do now?
		\item What do we show?
		\item What's the impact?
	\end{itemize}
}
}

\begin{abstract}
Rendering is the process of generating 2D images from 3D assets, simulated in a virtual environment, typically with a graphics pipeline. By inverting such renderer, one can think of a learning approach to predict a 3D shape from an input image. However, standard rendering pipelines involve a fundamental discretization step called rasterization, which prevents the rendering process to be differentiable, hence able to be learned. We present the first non-parametric and truly differentiable rasterizer based on silhouettes. Our method enables unsupervised learning for high-quality 3D mesh reconstruction from a single image. We call our framework ``soft rasterizer" as it provides an accurate soft approximation of the standard rasterizer. The key idea is to fuse the probabilistic contributions of all mesh triangles with respect to the rendered pixels. When combined with a mesh generator in a deep neural network, our soft rasterizer is able to generate an approximated silhouette of the generated polygon mesh in the forward pass. The rendering loss is back-propagated to supervise the mesh generation without the need of 3D training data. Experimental results demonstrate that our approach significantly outperforms the state-of-the-art unsupervised techniques, both quantitatively and qualitatively. We also show that our soft rasterizer can achieve comparable results to the cutting-edge supervised learning method~\cite{wang2018pixel2mesh} and in various cases even better ones, especially for real-world data.
\end{abstract}

\section{Introduction}
\label{sec:intro}

\nothing{
\weikai{Our advantages:
	\begin{itemize}
		\item Unbiased estimation of rasterization. Enable training in both forward and backward directions.
		\item Unsupervised training without the need of 3D groundtruth. Reduce a huge amount of labeling effort. 
		\item Novel approach of color reconstruction without use of the texture map. Can be applied to arbitrary genus-0 surface.
	\end{itemize}
}	
}

\nothing{
\note{Why differentiable renderer is important?
\begin{itemize}
	\item Image based 3D reconstruction problem is important
	\item Supervised learning requires huge amount of 3D ground truth, which is difficult to obtain. The key to unsupervised learning is differentiable rendering
\end{itemize}	
}
}

\begin{figure}[h!]
  	\centering
	\includegraphics[width=1\linewidth]{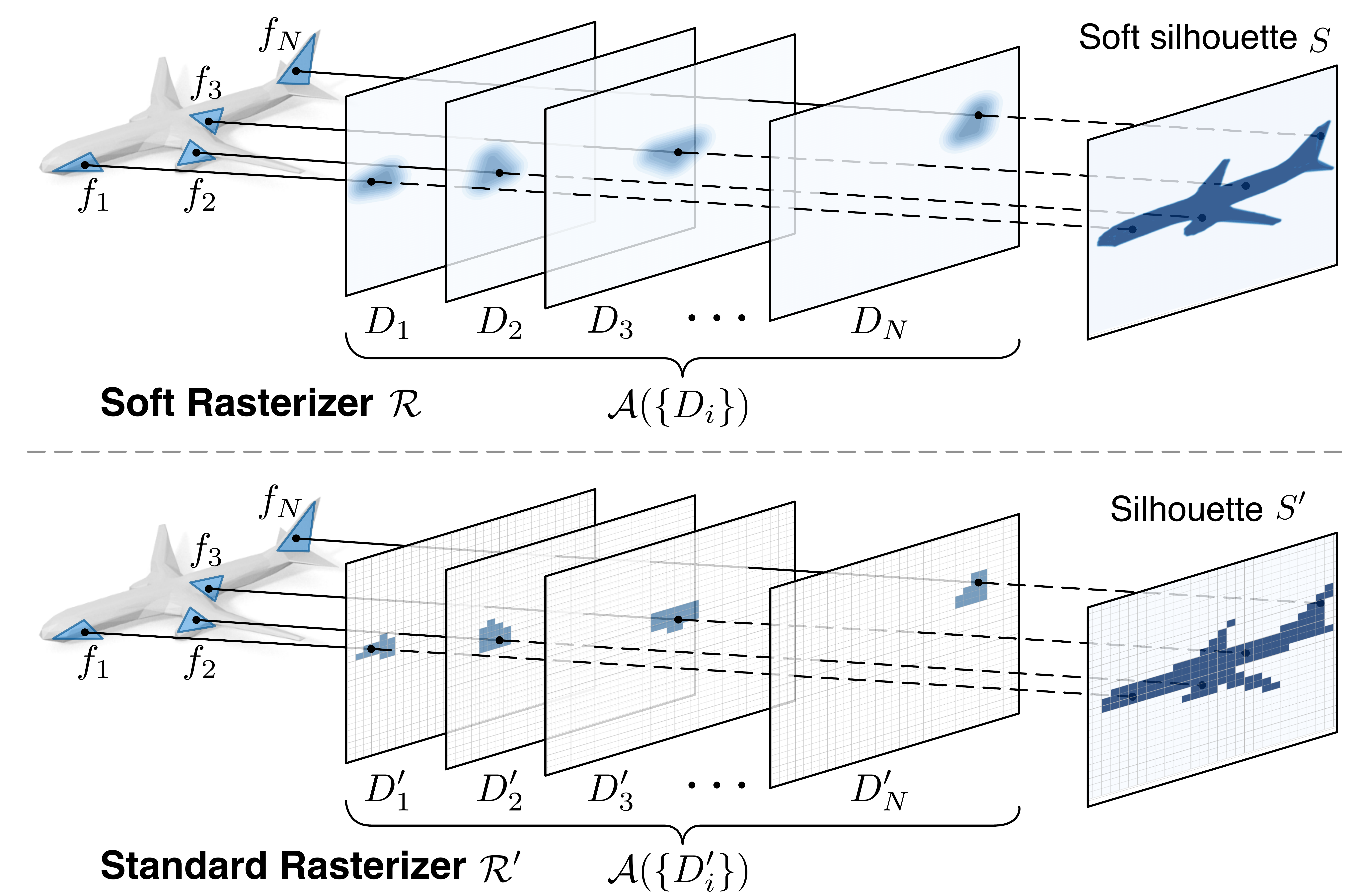}
\caption{Our differentiable Soft Rasterizer $\mathcal{R}$ (upper) can render mesh silhouette that faithfully approximates that generated by a standard rasterizer $\mathcal{R}'$ (below). 
$\mathcal{R}'$ renders a pixel as solid once it is covered by a projected triangle, leading to a discrete and non-differentiable process.
We propose to approximate the rasterized triangles $\{D_i'\}$ with a ``soft" continuous representation $\{D_i\}$ based on signed distance field.
We further fuse $\{D_i\}$ with a differentiable aggregate function $\aggregate(\cdot)$, which is essentially a logical {\it or} operator, so that the entire framework is differentiable. 
	}
  	\label{fig:teaser}
\end{figure}

\nothing{We propose the differentiable Soft Rasterizer (upper) to approximate the standard rasterizer (below) in graphics pipeline. We decompose rasterization into: 1) triangle-wise regions $\{\overline{D}_i\}$ in camera coordinate; 2) aggregate function $A(\cdot)$ that takes all regions to produce the raster image. In this perspective, we soften the hard triangle regions with signed distance maps $\{D_i\}$ so that the whole process will be differentiable. The proposed rasterizer could be integrated into neural networks.}

\begin{figure*}
	\centering

	\begin{subfigure}[b]{.13\linewidth}
		\centering
		\includegraphics[width=\linewidth,trim=0 30 0 30,clip]{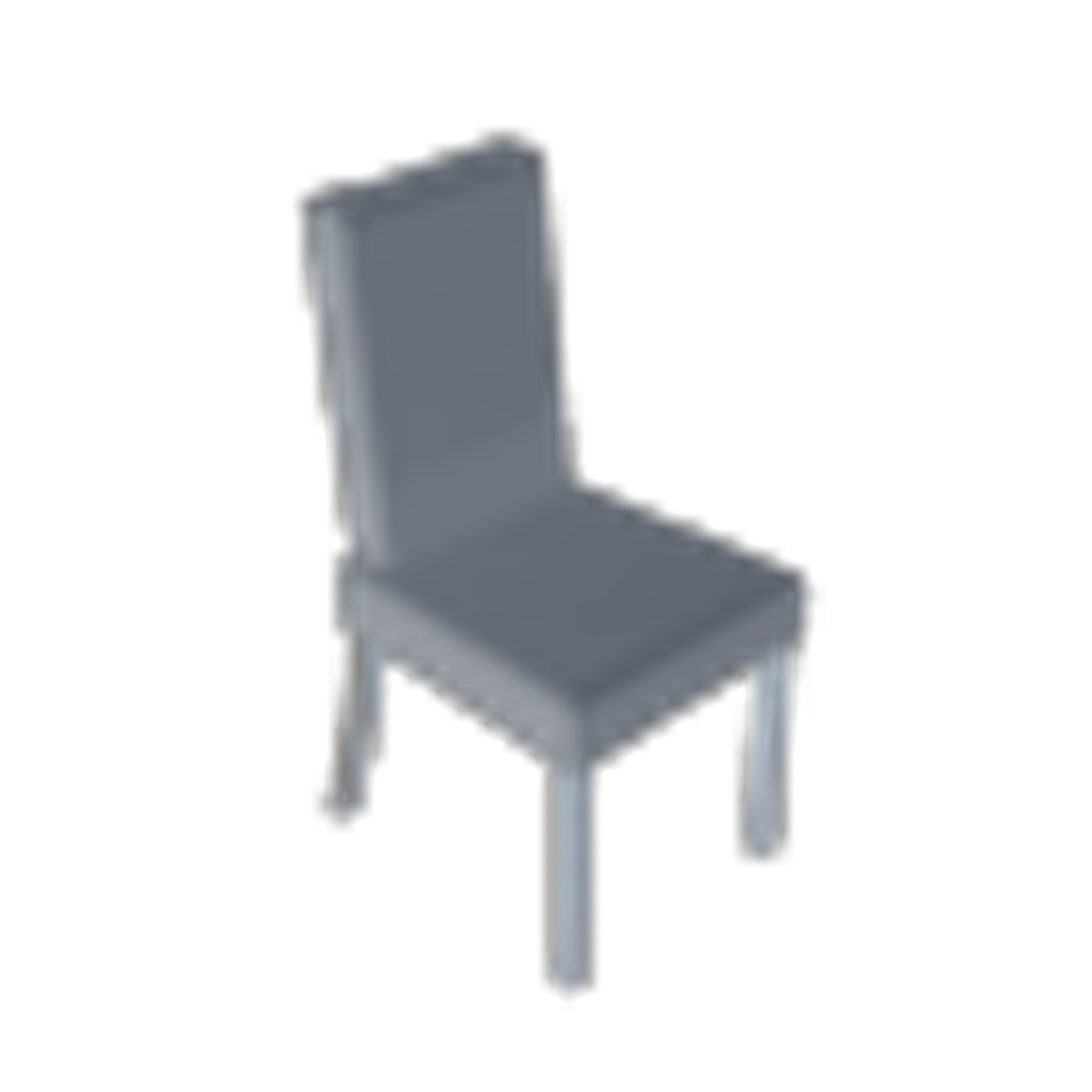}
			\caption{{\footnotesize Synthetic Image}}
	\end{subfigure}
	\begin{subfigure}[b]{.24\linewidth}
		\centering
		\includegraphics[width=\linewidth,trim=0 30 0 30,clip]{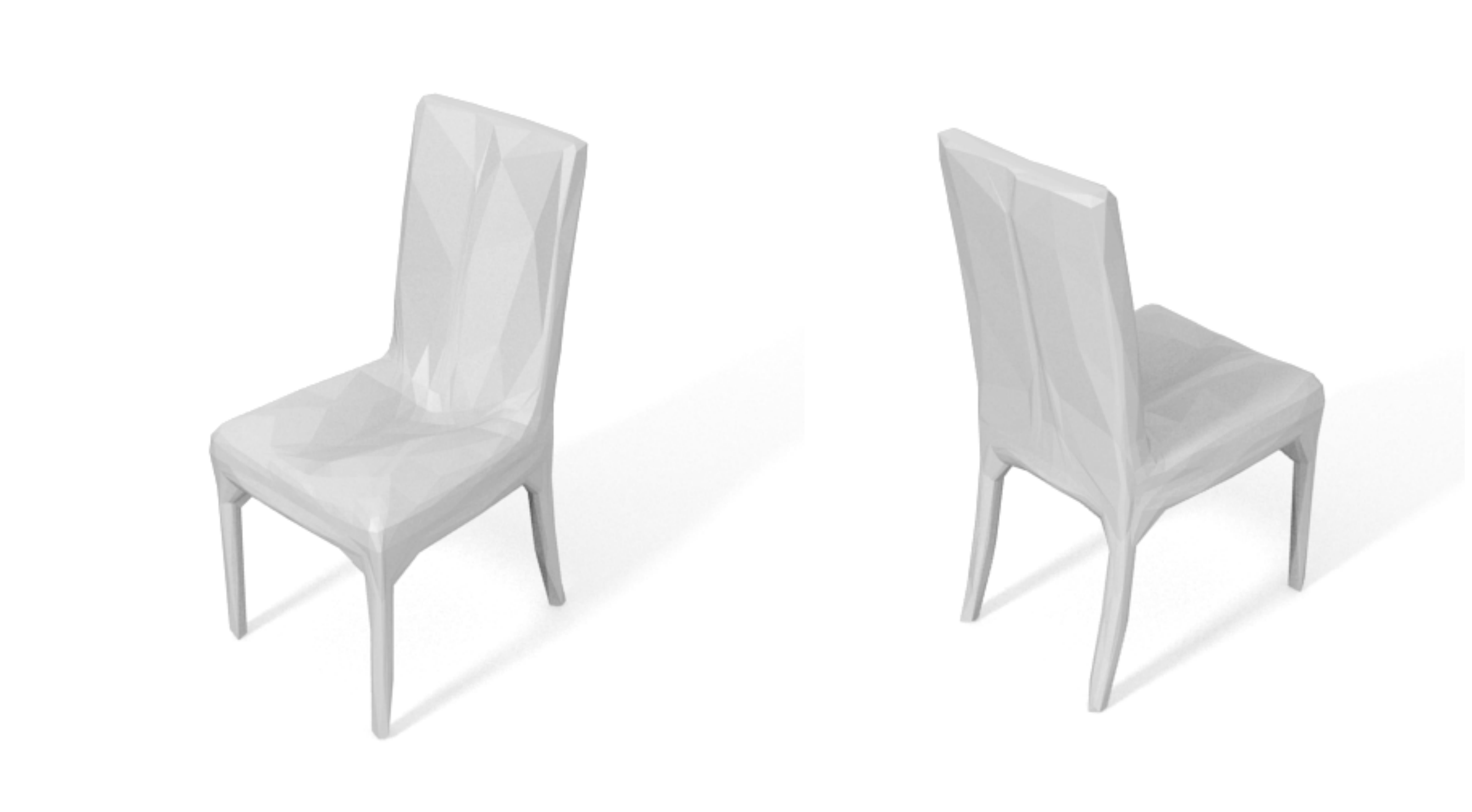}
		\caption{Our reconstruction}
	\end{subfigure}
	\begin{subfigure}[b]{.15\linewidth}
		\centering
		\includegraphics[width=\linewidth,trim=0 66 0 66,clip]{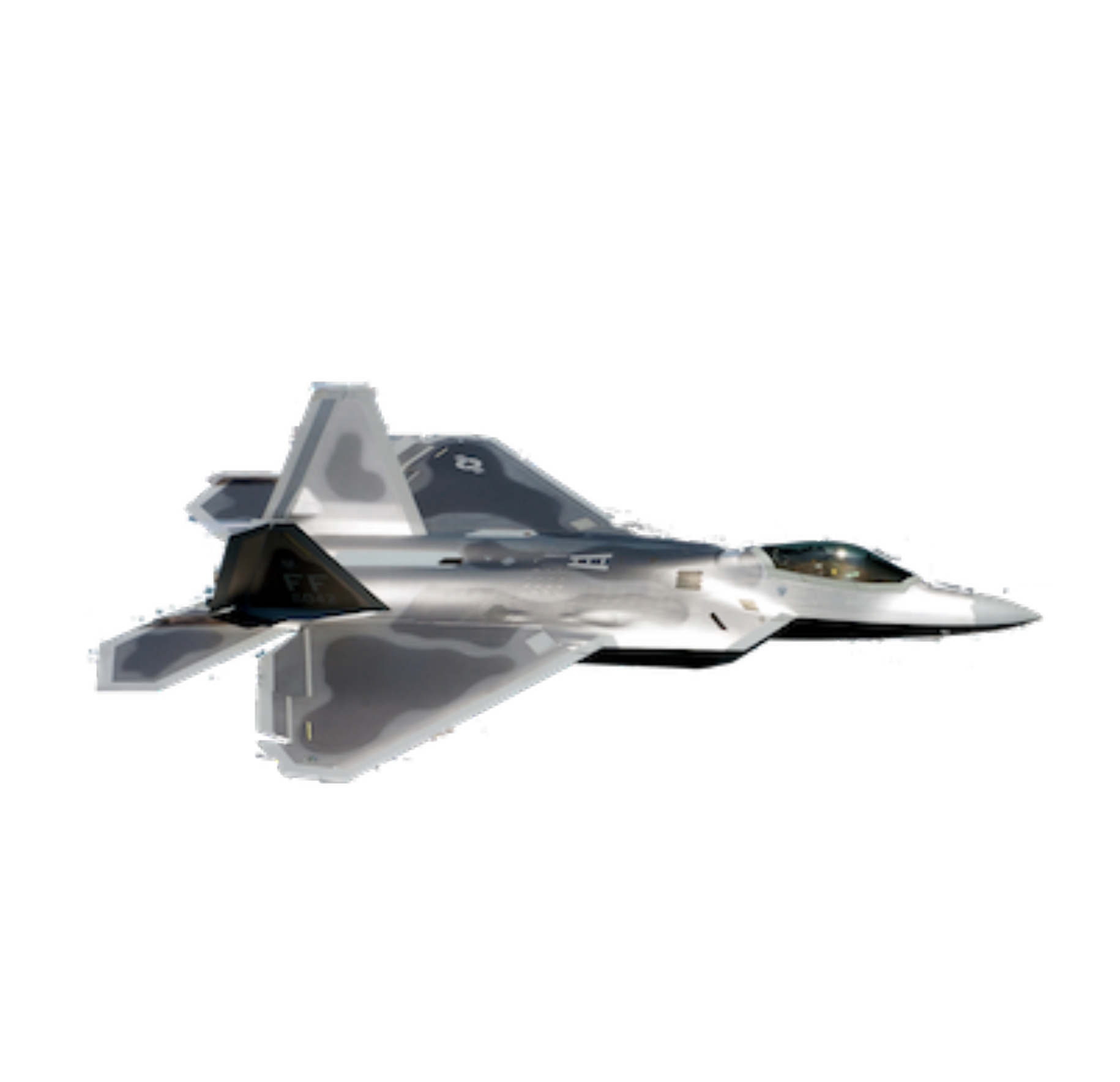}
		\caption{Real image}
	\end{subfigure}
	\begin{subfigure}[b]{.26\linewidth}
		\centering
		\includegraphics[width=\linewidth,trim=0 66 0 66,clip]{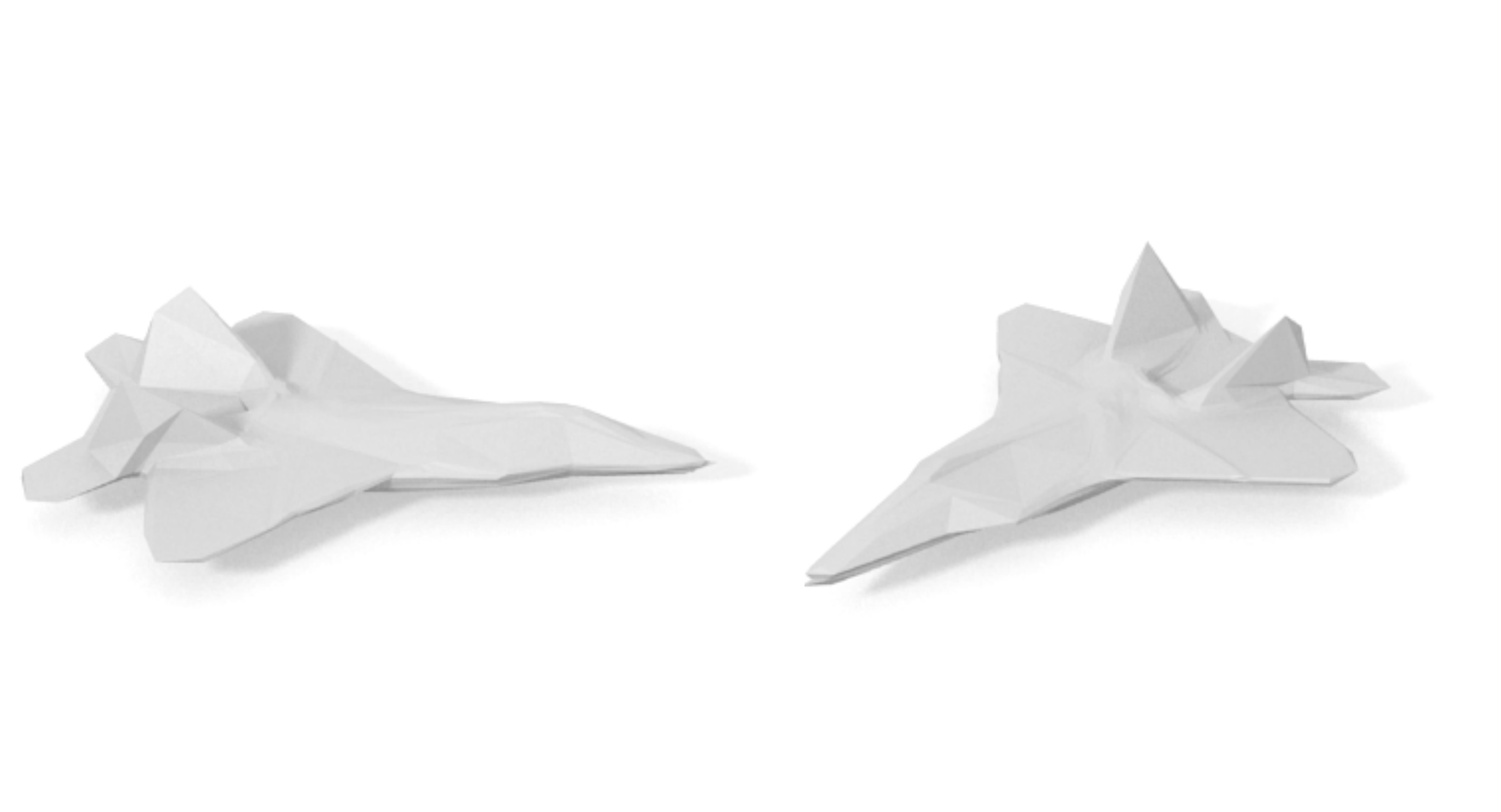}
		\caption{Our reconstruction}
	\end{subfigure}

	\vspace{-6px}
	\caption{Example reconstruction results using our approach on testing synthetic image and real image. }
	\label{fig:teaser2}
\end{figure*}

We live in a complex three-dimensional world, consisting of limitless 3D shapes of objects and matter, and
yet, our observations are only 2D projections of this world. One of the fundamental goals in computer vision, dating back to the sixties~\cite{roberts1963machine}, has been to build a computational system that can understand and reconstruct any 3D scenes, structures, and objects given a picture. Early attempts for single-view 3D modeling relied on hand-designed priors \cite{lowe1987three} or statistical models that describe the image formation process \cite{marr1982vision}. Later approaches include data-driven techniques that learn models from a collection of 3D data sets~\cite{hoiem2005automatic, saxena2009make3d}.

Recent advancements of deep learning have shown that the computational gap between 2D images and scene analysis is closing for a wide range of end-to-end tasks using supervised learning, such as image recognition \cite{ he2016deep}, object detection \cite{ren2015faster} and segmentation \cite{lin2016efficient}, {\it etc}. 
However, extending the paradigm of supervised learning for 3D inference is non trivial, due to the limited availability of 3D assets compared to 2D images, as well as the complexity of rendering all possibilities w.r.t. shape, texture, lighting exhaustively.
Consequentially, the ability to learn 3D deep models successfully without ground-truth supervision can lead to unprecedented possibilities for the general task of single-view 3D reconstruction tasks.

The key to unsupervised 3D inference is to find a way to relate the changes of non-domain specific 3D model parameters (e.g., geometry, illumination, material properties, {\it etc}) with those in the observed image. 
Differentiable rendering relates the derivatives of pixel intensities with the properties of the virtual object such that a 3D inference problem can be formulated by a gradient-based optimization without the need of supervised learning.

\nothing{
\note{What's the problems with previous works?}
}

However, the rendering procedure is not differentiable in conventional graphics pipelines.
In particular a 3D object is projected onto the 2D screen through a discretization step called {\it rasterization}, which is not differentiable, while the model projection itself is.

To enable unsupervised training for image-based reconstruction,
a large body of work \cite{ richardson2017learning, tewari2017mofa, tewari2018self, kundu20183d, genova2018unsupervised} have suggested various ways of approximating the rendering gradient computation in the backward pass. 
Most of the prior frameworks are designed for domain-specific purpose and remain difficult to handle more general cases.
Recently, Kato et al.~\cite{kato2018neural} have introduced the first general-purpose neural mesh renderer based on deep learning.
In \cite{kato2018neural}, they approximate the rasterization gradient with a hand-designed function which response is piecewise linear with respect to the displacement of vertices.
While promising results were shown, the proposed linear function is too simplified to fully model the nonlinearity of a real rasterizer.
Furthermore, their framework only approximates the backward gradient computation while directly using the standard rasterizer in the forward pass.
The inconsistency between the forward and backward propagations makes it difficult to fully exploit the effectiveness of the rendering layer. 

\nothing{
\note{Our objective. Technical challenge. Proposed approach. }
}

To address these issues, we propose the first {\it truly differentiable rasterizer} which is able to faithfully approximate the discrete rasterization in the forward pass of a deep neural network.
Given a polygon mesh, our rasterizer can directly generate an approximated silhouette of the input under a given view (Figure~\ref{fig:teaser} upper).
Our network is only trained with multi-view silhouettes.
The difference between our rendered result and the true silhouette can be back propagated to the mesh generator, enabling unsupervised learning without 3D training data.
At test time, our approach can reconstruct a high-quality 3D mesh from a single image. 

The key insight of our work is that we show how to formulate the deterministic sampling operation of a rasterizer as a probabilistic procedure.
While the standard rasterizer directly picks the color of the closest triangle in the viewing direction (Figure~\ref{fig:teaser} below), we propose that all triangles have contributions to each rendered pixel with a certain probability.
In particular, triangles which are closer to the projected pixels in screen space are more likely to be sampled from that pixel.
To this end, we approximate the probabilistic contribution of each triangle face to the rendered pixels as a normalized distance field computed on the image plane (Figure~\ref{fig:teaser} upper). 
By passing the collected distance field to a differentiable aggregate function $\aggregate(\cdot)$, which simulates the logical {\it or} operator, our rasterizer can directly generate an approximated silhouette of the mesh.
We call our framework {\it Soft Rasterizer (SoftRas)} as it is a soft approximation of the standard rasterizer.

SoftRas itself does not involve any trainable parameters and thus can serve as a flexible module for other applications.
In addition, as we target for mesh reconstruction, our approach is able to generate higher quality results with much lower computational cost compared to techniques with voxel or point cloud based representations. 
Figure~\ref{fig:teaser2} shows examples of our model reconstructing a chair and fighter aircraft from a single image.
We also show that our approach significantly outperforms existing unsupervised methods w.r.t. quantitative and qualitative measures.
Furthermore, our experimental results indicate that our method can achieve comparable and in certain cases, even superior results to supervised solutions \cite{wang2018pixel2mesh}, indicating the effectiveness of the soft rasterizer framework.





\nothing{
\note{What are the advantages of our approach over other methods? What's our main contributions?}
}

\section{Related Work}
\label{sec:related_work}

\paragraph{Differentiable Renderer.}

The standpoint of viewing vision problems as {\it inverse graphics} has been investigated since the very beginning of the field \cite{baumgart1974geometric, yu1999inverse, patow2003survey}.
Through inverting the rendering progress, inverse graphics aims to infer the object shape, illumination and reflectance from an image. 
To relate the changes in the observed image with that in the 3D model parameters, there are a number of existing techniques utilizing the derivatives of rendering.

Gkioulekas et al.~\cite{gkioulekas2013inverse} build a material dictionary and  propose a direct optimization framework to invert volumetric scattering using stochastic gradient descent.
In \cite{gkioulekas2016evaluation}, researchers present an analytical formulation between the measurements and internal scattering parameters which enables a derivative-based optimization. 
Though the gradients are leveraged for solving the inverse problems, these approaches are limited to specific light transporting problems. 
Mansinghka et al. \cite{mansinghka2013approximate} propose a general inverse rendering technique by using a probabilistic graphics model to infer scene parameters from observations.
More recently, Loper and Black~\cite{loper2014opendr} further introduce OpenDR, an approximate differentiable renderer which can be incorporated into probabilistic programming framework to obtain derivatives with respect to the model parameters.

With the recent surge of convolutional neural network, there is an increasing popularity to consider rendering derivatives in a deep learning framework.
In particular, many learning-based techniques \cite{zienkiewicz2016real, liu2017material, richardson2017learning, tewari2017mofa, tewari2018self, deschaintre2018single, kundu20183d, nguyen2018rendernet, genova2018unsupervised} have incorporated a {\it differentiable rendering layer} to enable an end-to-end architecture for 3D reconstruction and material inference in an unsupervised manner.
However, these rendering layers are usually designed for special purpose and thus cannot be generalized to more general cases. 
Nalbach et al.~\cite{nalbach2017deep} propose a general deep shading network which learns the direct mapping from a variety of deferred shading buffers to corresponding shaded images.
Rezende et al.~\cite{rezende2016unsupervised} pioneer in unsupervised 3D structure generation from a single image using a differentiable renderer.
Later, a differentiable rendering pipeline specialized for mesh reconstruction was proposed by Kato et al.~\cite{kato2018neural} to approximate the gradient of pixel intensity with respect to mesh vertices.
By using \cite{kato2018neural}, Kanazawa et al.~\cite{kanazawa2018learning} strive to reconstruct category-specific mesh from image collections without relying on 3D ground-truth models. 
Apart from rasterization-based rendering, Li et al.~\cite{Li:2018:DMC} introduce a differentiable ray tracer to realize secondary rendering effects in a deep neural network.

In this paper, we focus on exploring a general differentiable framework for rasterization-based rendering. 
Unlike Neural 3D Mesh Renderer \cite{kato2018neural}, which approximates the discrete rasterization operation with a straightforward linear function, 
our approach is capable to provide the estimation of rendering derivative with significantly higher accuracy. 

\paragraph{Single-view 3D reconstruction.}

Image based 3D reconstruction is a long-standing problem in computer vision. 
Single image based reconstruction problem is especially challenging due to the mismatch between the scarcity of input and the redundancy of the plausible solutions.
Recent advances in machine learning address this issue by learning the priors of 3D properties, such as shape, reflectance or illumination \cite{barron2015shape, blanz1999morphable, shi2017learning} to reduce the searching space. 
Before the advent of differentiable renderer, a majority of prior works use supervised deep learning approaches, which seek to learn the nonlinear mapping between the input image and 3D model parameters \cite{tran2018nonlinear, yamaguchi2018high, zhou2018single, sengupta2018sfsnet, wang2018pixel2mesh} from labeled ground-truth data.
To simplify the learning problem, some works reconstruct 3D shape via predicting intermediate 2.5D representations, such as depth map \cite{ shelhamer2015scene, eigen2015predicting, liu2016learning}, visual hull \cite{natsume2018siclope}, implicit field \cite{huang2018deep}, displacement map \cite{huynh2018mesoscopic} or normal map \cite{bansal2016marr, qi2018geonet, wang2015designing}. 
When considering reconstructing 3D shape, voxel-based representation has received most attention \cite{wu2017marrnet, zhu2017rethinking, tulsiani2017multi} due to its simplicity of regular structure and compatibility with convolutional neural network. 
However, volumetric representation is constrained by its resolution due to the data sparsity and high computational cost.
Hence, recent progress on supervised learning has explored the avenue of reconstructing mesh \cite{wang2018pixel2mesh, groueix2018, pumarola2018geometry} or point cloud \cite{fan2017point} directly from a single image.

Comparing to supervised learning, unsupervised 3D reconstruction is becoming increasingly important as collecting ground-truth 3D models is much more difficult than labeling 2D images.
Perspective transformer nets~\cite{yan2016perspective} propose an encoder-decoder network which learns 3D shape from silhouette images in an unsupervised fashion. 
The key to their approach is adding a projection transformation as a self-supervised loss for regularizing the voxel generation.
Though our geometry reconstruction is also based on silhouette images, we use mesh representation which is much more computational efficient and can reconstruct geometry with much higher precision compared to the volumetric representation in \cite{yan2016perspective}.

\nothing{
\weikai{ add ref:
Geometry-Aware Network for Non-Rigid Shape Prediction from a Single View
\cite{pumarola2018geometry}
}
}

\section{Soft Rasterizer}
\label{sec:rasterizer}

\nothing{
\shichen{Better contains: (1) probability of a pixel in each single triangle ($D_i$), (2) probability of a pixel in all triangles (Aggregate $\{D_i\}$), (3) compare with previous methods. Use fig\_raster to illustrate (1) and (2).}
}

The main obstacle that impedes the standard graphics renderer from being differentiable is the discrete sampling operation, which is also named {\it rasterization}, that converts a continuous vector graphics into a raster image.
In particular, after projecting the mesh triangles onto the screen space, standard rasterization technique fills each pixel with the color from the nearest triangle which covers that pixel.
However, the color intensity of an image is the result of complex interplay between a variety of factors, including the lighting condition, viewing direction, reflectance property and the intrinsic texture of the rendered object, most of which are entirely independent from the 3D shape of the target object. 
Though one can infer fine surface details from the shading cues, special care has to be taken to decompose shading from the reflectance layers. 
Therefore, leveraging color information for 3D geometry reconstruction may unnecessarily complicate the problem especially when the target object only consists of smooth surfaces. 
As a pioneering attempt for reconstructing general objects, our work only focuses on synthesizing silhouettes, which are solely determined by the 3D geometry of the object.

\begin{figure}[h]
	\begin{subfigure}[b]{.32\linewidth}
    	\centering
    	\includegraphics[width=\linewidth]{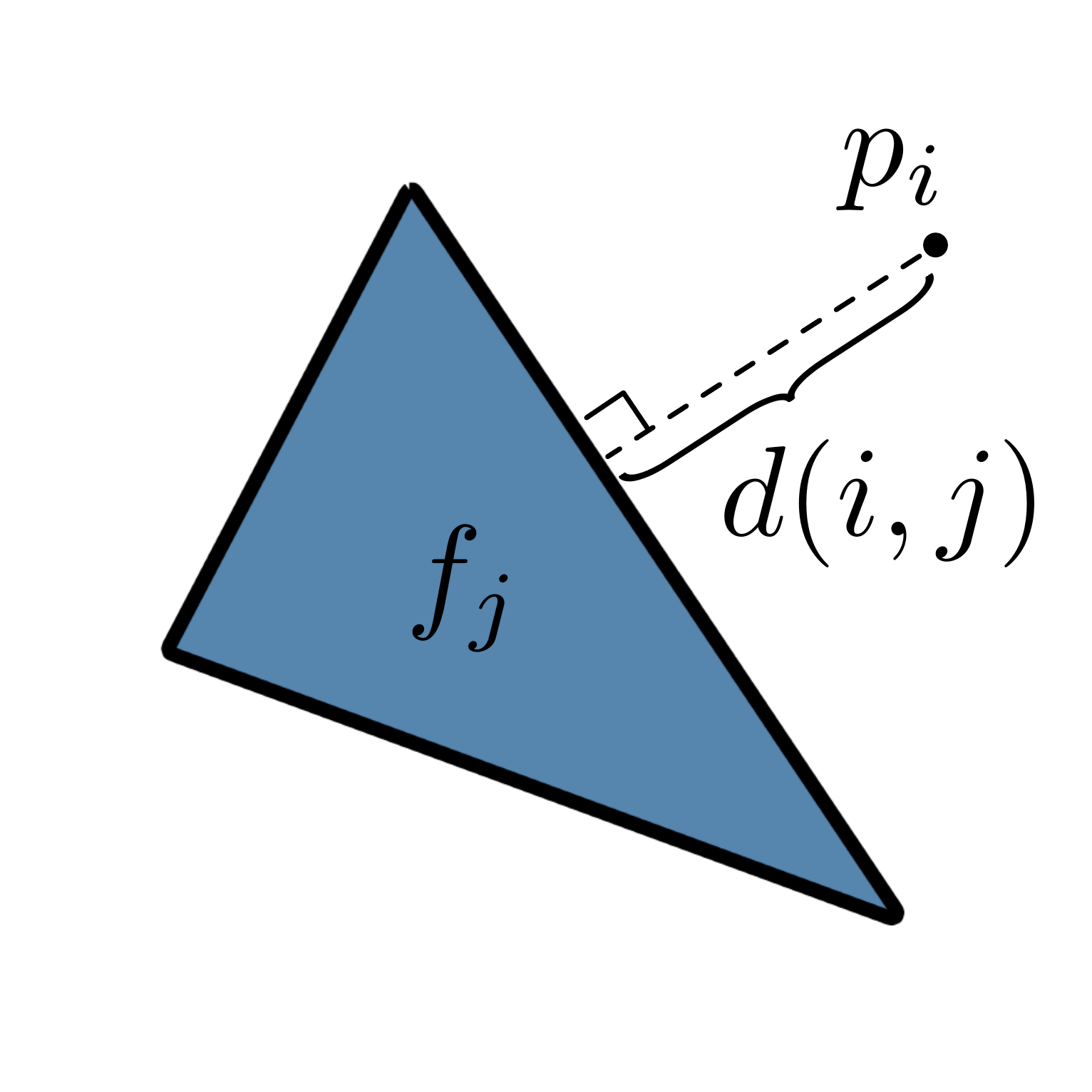}\\
    	\caption{ground truth}
	\end{subfigure}
	\begin{subfigure}[b]{.32\linewidth}
    	\centering
    	\includegraphics[width=\linewidth]{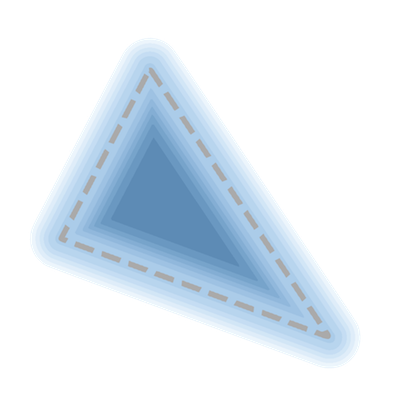}\\
    	\caption{$\sigma = 0.01$}
	\end{subfigure}
	\begin{subfigure}[b]{.32\linewidth}
    	\centering
    	\includegraphics[width=\linewidth]{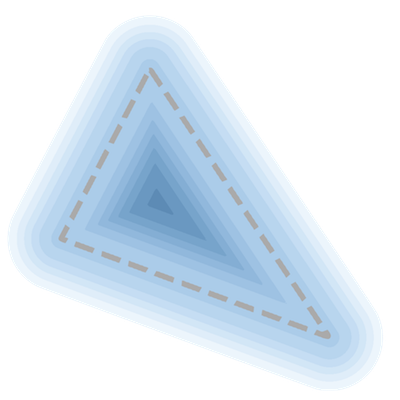}\\
    	\caption{$\sigma = 0.03$}
	\end{subfigure}

   	\caption{Example probability maps of a single triangle. (a): definition of pixel-to-triangle distance; (b) and (c): probability maps generated with different $\sigma$.}
   	\label{fig:triangle}

\end{figure}

\begin{figure*}[t]
	\centering
	\includegraphics[width=1\linewidth]{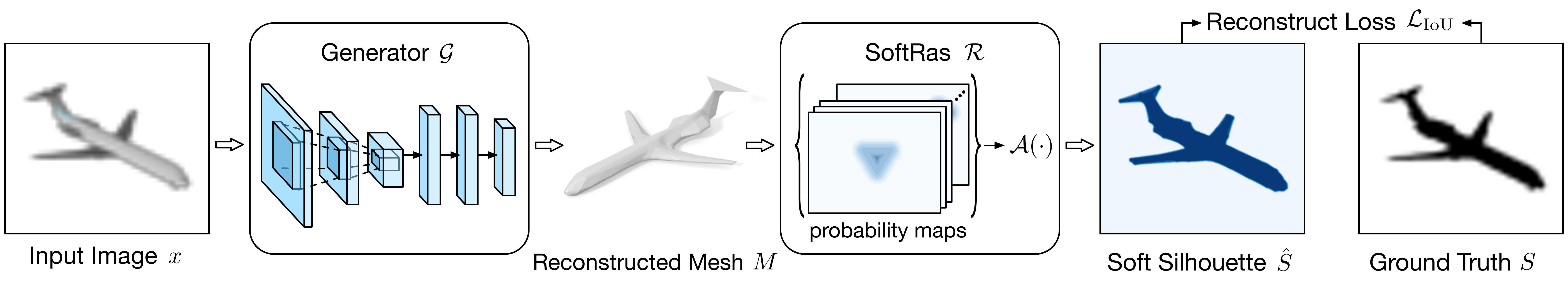}
	\caption{Our pipeline for unsupervised single-image 3D mesh reconstruction: (1) mesh generator $\mathcal{G}$ takes an input image $x$ and outputs the displacement vector $\mathcal{G}(x)$, which is added to a template model to obtain the reconstructed mesh $M$; (2) SoftRas layer $\mathcal{R}$ transforms triangles in $M$ into probability maps and computes soft silhouette $\hat{S}$ using the aggregation function $\mathcal{A}(\cdot)$. IoU loss is then applied to  minimize the discrepancy between $\hat{S}$ and ground truth $S$. 
		}
	  \label{fig:framework}
	  \vspace{-5px}
\end{figure*}

Unlike color image, the silhouette image is a binary mask where a pixel is labeled as solid as long as it is covered by any triangle. 
To differentiate such discrete operation, we propose to formulate it as a probabilistic procedure.
In particular, instead of viewing sampling as a deterministic event, we assume that each triangle face can potentially cover a specific pixel with certain probability.
Such probability distribution is highly related with the distance $\dist(i, j)$ between the specified pixel $\pixel_i$ and the triangle $\face_j$ (Figure~\ref{fig:triangle}).
The closer $\pixel_i$ is to the center of $\face_j$, the more likely $\pixel_i$ is covered by $\face_j$. 
Towards this end, we encode the probabilistic ``contribution" of triangle $\face_j$ to all image pixels in the form of {\it probability map}, which is denoted as $\Dmap_j$. 
The value at the $i$-th pixel of $\Dmap_j$ encodes the probability of $\face_j$ covers the corresponding pixel in the final rendered image.
After the probability maps $\{\Dmap_j\}$ of all triangles are obtained, we approximate the rasterization result by fusing $\{\Dmap_j\}$ with a specially-designed aggregation function $\aggregate(\cdot)$.
We describe the details of computing probability maps $\{\Dmap_j\}$ and aggregation function $\aggregate(\cdot)$ in the following sections.

\subsection{Probability Map Computation}  
\label{sec:prob_map}

The probability map $\Dmap_j$ of each face $\face_j$ has the same resolution ($\height \times \width$) with the output silhouette image $\sil$.
As discussed before, when computing $\Dmap_j$, the region that is closer to the triangle center, especially those enclosed by $\face_j$, shall receive higher weight in a properly designed probability distribution.
In contrast, the probability intensity of the pixels that are outside $\face_j$ should drop fast as their distance to $\face_j$ increases.
It is easy to observe that such probability distribution is closely related to the distance field with respect to the boundary of the triangle $\face_j$.
To this end, we propose the formulation of $\Dmap_j$ as follows:

\begin{equation}
\D_j^i = sigmoid(\sig_{ij} \cdot \frac{d^2(i,j)}{\sigma}),
\label{eqn:di}
\end{equation}

\nothing{
where $\D_j^i$ is the probability value at the $i$-th pixel $\pixel_i$ (scan line order) of $\Dmap_j$; $\dist(i,j)$ returns the shortest distance from $p_i$ to the edges of $\face_j$ (see Figure~\ref{fig:triangle}(a)); $\sigma$ is a positive hyperparameter that controls the sharpness of the probability distribution while $\sig_{ij}$ is a signed indicator whose response depends on the relative position between $\pixel_i$ and $\face_j$: 
where $\D_j^i$ is the probability value at the $i$-th pixel $\pixel_i$ (scan line order) of $\Dmap_j$; $\dist(i,j)$ returns the shortest distance from $p_i$ to $\face_i$; $\sigma$ is a positive hyperparameter that controls the sharpness of the probability distribution while $\sig{ij}$ is a signed indicator whose response depends on the relative position between $\pixel_i$ and $\face_j$: \shichen{could be more detailed}
}

where $\D_j^i$ is the probability value at the $i$-th pixel $\pixel_i$ (scan line order) of $\Dmap_j$; $\dist(i,j)$ returns the shortest distance from $p_i$ to the edges of $\face_j$ (see Figure~\ref{fig:triangle}(a)); $\sigma$ is a positive hyperparameter that controls the sharpness of the probability distribution while $\sig_{ij}$ is a signed indicator whose response depends on the relative position between $\pixel_i$ and $\face_j$:

\begin{align*}
	\sig_{ij} =
	\begin{cases}
	+1 & \mathrm{if} \ \pixel_i \in \face_j\\
	-1 & \mathrm{otherwise}.
	\end{cases}
\end{align*}

Intuitively, by using the {\it sigmoid} function, Equation~\ref{eqn:di} generates a normalized output staying within the interval $\left(0, 1\right)$.
In addition, the introduction of the signed indicator maps pixels inside and outside $\face_j$ to the probability distribution of $\left(0.5, 1\right)$ and $\left(0, 0.5 \right)$ respectively. 
Figure~\ref{fig:triangle} shows $\Dmap_j$ of a particular triangle with different configurations of $\sigma$.
As shown in the results, smaller $\sigma$ leads to sharper probability distribution while larger $\sigma$ tends to generate more blurry outcome.
As $\sigma\rightarrow 0$, the resulting probability field converges to the exact silhouette of the triangle.

It is worth noting that before computing the pixel-to-triangle distance $\dist(i,j)$, we first normalize the pixel coordinates into $\left[-1, 1\right]$.
This helps factor out the bias introduced by different image resolutions.
Though it is possible to directly use $\dist(i,j)$ in the computation of $\Dmap_j$, in practice, we found $\dist^2(i,j)$ works better as it introduces a sharper distribution of the probability field.

\subsection{Aggregate Function}
\label{sec:aggregate}

As the probability map $\Dmap_j$ of each triangle face is computed in an independent manner, there remains a key step to aggregate them to obtain the final rendering result. 
To achieve this goal, we first analyze the working principal of standard rendering pipeline.
The standard rasterizer synthesizes silhouette image by following a strict logical {\it or} operation.
In particular, a pixel that is covered by any triangle on the image plane will be rendered as the interior of the object in the output silhouette image.
As our proposed probability formulation has emphasized the inner pixels enclosed by each triangle with higher weight -- their probability values are closer to 1 -- we approximate the logical {\it or} operator by proposing the following aggregation function:

\begin{equation}
\silSmall^i = \aggregate(\{\Dmap_j\}) = 1- \prod_j^N (1 - \D_j^i),	
\end{equation}

where $\silSmall^i$ represents the  $i$-th pixel in the final silhouette image and $N$ is the number of total mesh triangles.
Intuitively, when there exists a $\D_j^i$ whose value is 1, then $\silSmall^i$ will be marked as 1 regardless the probability value of other probability maps at this pixel location.
On the other hand, $\silSmall^j$ will receive 0 only when all $\{ \D_j^i \}$ are zeros.

Figure~\ref{fig:framework} demonstrates the generated silhouette using our \model \ with $\sigma = 3\times10^{-5}$. 
As shown in the result, \modelshort is able to faithfully approximate the ground-truth silhouette without losing fine details.
However, unlike the standard rasterizer, which includes discrete non-differentiable operations, our \modelshort is entirely differentiable thanks to its continuous definition.

\section{Unsupervised Mesh Reconstruction}
\label{sec:overview}

To demonstrate the effectiveness of the proposed \model, we incorporate it into a simple mesh generator for the task of image-based 3D reconstruction.
In particular, we propose an end-to-end deep learning framework that takes a single color image $\inImg$ as input and reconstructs its corresponding 3D mesh model $\mesh$ without requiring the camera parameters. 
As illustrated in the framework overview in Figure~\ref{fig:framework}, our system mainly consists of two parts: one mesh generator that deforms a template sphere mesh to the desired 3D model and a soft rasterizer layer that renders the  silhouette image of the reconstructed mesh.

\subsection{Mesh Generator}
\label{sec:meshGen}

Inspired by the state-of-the-art mesh learning approaches \cite{kato2018neural,wang2018pixel2mesh}, we leverage a similar idea of synthesizing 3D model by deforming a template mesh.
To validate the performance of our proposed \model, we employ an encoder-decoder architecture which is nearly identical to that of \cite{kato2018neural}.
In particular, the mesh generator consumes a 4-channel color image, in which the last channel specifies the object's silhouette mask, and predicts per-vertex displacement vectors that deform the template mesh to the desired shape.
We use the same template mesh -- a sphere model -- with Neural 3D Mesh Renderer \cite{kato2018neural}. 

The training of our network does not require any 3D ground truth.
Specifically, our network is only trained with multi-view silhouettes of the objects collected in the training set.
Thanks to the differentiability of \model, the learning of mesh generator is directly supervised by the rendering loss computed from the difference between
the rendered and the ground-truth silhouettes.

\begin{figure*}[ht!]
    \newlength{\myseqwidth}
    \setlength{\myseqwidth}{.136\linewidth}
    \vspace{-10px}
	\begin{center}

        \includegraphics[width=1.0\linewidth]{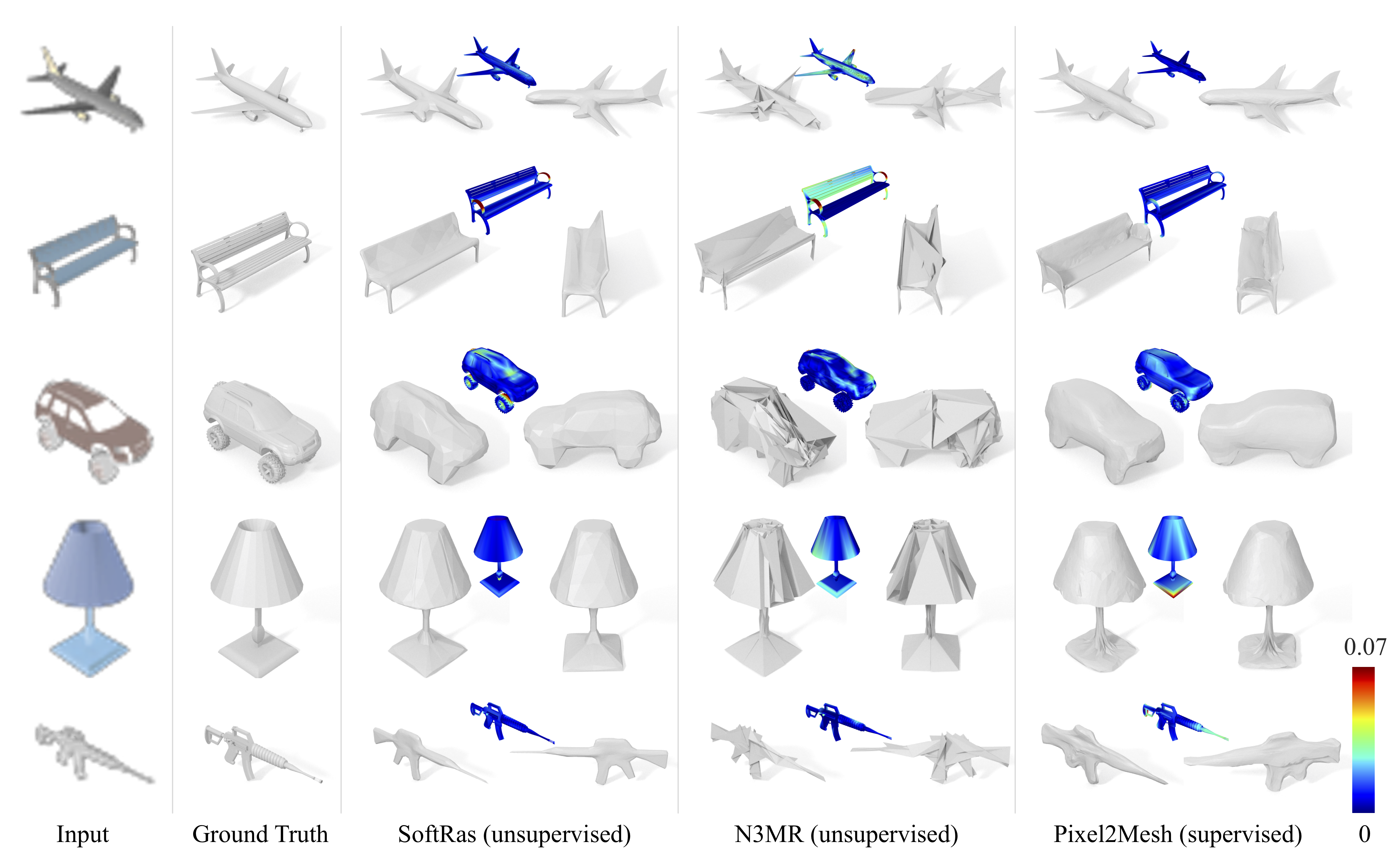}

   \end{center}
   \vspace{-8px}
   \caption{3D mesh reconstruction from a single image. From left to right, we show input image, ground truth, the results of our method (SoftRas), Neural Mesh Renderer~\cite{kato2018neural} and Pixel2mesh~\cite{wang2018pixel2mesh}, all visualized from 2 different views. Along with the results, we also visualize scan-to-mesh distances measured from ground truth to reconstructed mesh.
   }
   \label{fig:vis_compare}
\end{figure*}

\begin{table*}[h!]
\addtolength{\tabcolsep}{4pt}
\centering
\begin{tabular}{ccccccccc}
\hline
Category    & Airplane & Bench  & Dresser & Car    & Chair  & Display & Lamp   &        \\ 
\hline
retrieval~\cite{yan2016perspective}   & 0.5564   & 0.4875 & 0.5713  & 0.6519 & 0.3512 & 0.3958  & 0.2905 &        \\
voxel-based~\cite{yan2016perspective} & 0.5556   & 0.4924 & 0.6823  & 0.7123 & 0.4494 & 0.5395  & 0.4223 &        \\
N3MR~\cite{kato2018neural}        & 0.6172   & 0.4998 & \bf{0.7143}  & 0.7095 & 0.4990 & 0.5831  & 0.4126 &        \\
\modelshort (ours)     & \bf{0.6419}   & \bf{0.5080} & 0.7116  & \bf{0.7697} & \bf{0.5270} & \bf{0.6156}  & \bf{0.4628} &        \\ 
\hline
Category    & Loudspeaker  & Rifle  & Sofa    & Table  & Phone  & Vessel  &        & Mean   \\ \hline
retrieval~\cite{yan2016perspective}   & 0.4600   & 0.5133 & 0.5314  & 0.3097 & 0.6696 & 0.4078  &        & 0.4766 \\
voxel-based~\cite{yan2016perspective} & 0.5868   & 0.5987 & 0.6221  & \bf{0.4938} & 0.7504 & 0.5507  &        & 0.5736 \\
N3MR~\cite{kato2018neural}        & 0.6536   & 0.6322 & 0.6735  & 0.4829 & 0.7777 & 0.5645  &        & 0.6015 \\
\modelshort (ours)     & \bf{0.6654}   & \bf{0.6811} & \bf{0.6878}  & 0.4487 & \bf{0.7895} & \bf{0.5953}  &        & \bf{0.6234} \\ 
\hline
\end{tabular}

\vspace{2pt}
\caption{Comparison of mean IoU with other unsupervised 3D reconstruction methods on 13 categories of ShapeNet datasets.}
\label{tab:shapenet}
\vspace{-5mm}
\end{table*}

\paragraph{Losses.} As object silhouette is represented in the form of binary mask, to evaluate the accuracy of our prediction, we adopt the intersection over union (IoU) loss $\loss_{IoU}$ for a proper measurement. 
In addition, to enforce the generator produces smooth and appealing results, we impose two additional geometry regularizer to constrain the property of output shape.
We provide the details of losses as follows.
   
{\it IoU Loss.} We denote $\sil$ and $\silGT$ as the binary masks of the reconstructed and ground-truth silhouette respectively and define $\otimes$ and $\oplus$ be the operators that performs element-wise product and sum respectively.
Therefore our IoU loss can be represented as:

\begin{equation}
	\loss_{IoU} = 1 - \frac{||\sil \otimes \silGT||_1}{||\sil \oplus \silGT - \sil \otimes \silGT||_1}
\end{equation}

{\it Laplacian Loss.} A simple IoU loss only focuses on pushing the mesh projection to be consistent with the true silhouette but could lead to strongly deformed mesh due to the priority to favor local consistency.
To prevent the vertices from moving too freely, we add a Laplacian term to regularize the geometry.
Let $\mesh$ be the output triangular mesh with $n$ vertices.
Each vertex $i \in \mesh$ is denoted as $\vertex_i = (x_i, y_i, z_i)$. 
We first define the laplacian coordinate for $\vertex_i$ as the difference between the coordinates of $\vertex_i$ and the center of mass of its immediate neighbors:
$\delta_i = \vertex_i - \frac{1}{||N(i)||}\sum_{j\in N(i)} \vertex_j$.
The laplacian loss is defined as:

\begin{equation}
	\loss_{lap} = \sum_i ||\delta_i ||^2_2
\end{equation}

{\it Flattening Loss.} In addition to laplacian loss, we also employ a flattening loss \cite{kato2018neural} to encourage adjacent triangle faces to have similar normal directions. 
Empirically, we found the introduced flattening loss can further smooth the surface and prevent self-intersections.
To calculate the flattening loss, we set $\theta_i$ to be the angle between the faces that have the common edge $e_i$.
Therefore, the flattening loss can be defined as:

\begin{equation}
	\loss_{fl} = \sum_{\theta_i \in {e_i}}{(\cos \theta_i + 1)^2}
\end{equation}

where $\loss_{fl}$ will reach its minimum value if all adjacent faces stay on the same plane.
The final loss is a weighted sum of the three particular losses:

\begin{equation}
	\loss = \loss_{IoU} + \lambda \loss_{lap} + \mu \loss_{fl}
	\label{equ:loss_all}
\end{equation}

\paragraph{Color Reconstruction.}
\label{sec:color}

Unlike vertex position, the vertex color can naturally receive gradients back propagated from the image loss.
Thanks to the proposed differentiable rasterizer, we are able to recover both 3D geometry and the accompanied vertex color in an end-to-end manner.  
In particular, we leverage the $l_2$ loss to measure the difference between the projection of recovered colorful mesh and the input image.
We show the results of colorized reconstruction in Section~\ref{sec:quali_results}. 




\section{Experiments}
\label{sec:result}
\vspace{-2mm}
In this section, we perform an extensive evaluation on our framework. 
We first provide the details of our experimental setups and then demonstrate a variety of results and ablation studies.
We also include video result and more visual evaluations in the \textit{supplemental materials}.
The code and data will be released upon publication.

\subsection{Experimental Setup}
\label{sec:setup}

\paragraph{Datasets.} 
We use the dataset provided by \cite{kato2018neural}.
In particular, the dataset contains 13 categories of objects belonging to ShapeNet~\cite{chang2015shapenet}, a large-scale 3D model collection containing about 50k unique meshes from 55 categories.
Each object is rendered with 24 different views with image resolution of 64 $\times$ 64. 
To ensure fair comparison, we employ the same train/validate/test split as in \cite{kato2018neural, yan2016perspective}. 

\paragraph{Evaluation Metrics.} 
For quantitative evaluation, we adopt the standard reconstruction metric, 3D intersection over union (IoU), to compare with baseline methods.  
On the other hand, we agree with Pixel2Mesh~\cite{wang2018pixel2mesh} that the commonly used metric may not be able to thoroughly reflect the quality of geometry reconstruction, such as smoothness and continuity of the surface.
Therefore, we wish to emphasize the importance of visual quality, which is important in real applications.


\paragraph{Implementation Details.} 
We use similar mesh generation network with \cite{kato2018neural}, which is implemented as an encoder-decoder architecture.
The encoder netowrk consists of 3 convolutional layers with kernel size of 5 $\times$ 5 and channels of 64, 128, 256, followed by 3 fully connected layers with hidden layer size of 1024 and 1024.
In particular, we apply batch normalization \cite{ioffe2015batch} and ReLU activation \cite{krizhevsky2012imagenet} after each convolutional layer.
The decoder transforms a 512-dimensional latent code into a displacement vector of length 1926 (coordinates of 642 vertices) by using 3 fully connected layers with hidden size of 1024 and 2048. 
Our network is trained with Adam optimizer~\cite{kingma2014adam} with $\alpha = 0.0001, \beta_1 = 0.9$ and $\beta_2 = 0.999$. 
In our implementation, we set $\sigma = 3\times 10^{-5}$ (Equation~\ref{eqn:di}), $\lambda = 0.01$ and $\mu = 0.001$ (Equation~\ref{equ:loss_all}) across all experiments unless otherwise specified. 
We train the network with batch size of 64 with multi-view inputs and implement it using PyTorch\footnote{\url{https://pytorch.org/}}. 
The code and data will be released upon publication.


\begin{figure}[t]
	\begin{subfigure}[b]{.24\linewidth}
		\centering
		\includegraphics[width=\linewidth,trim=30 30 30 30,clip]{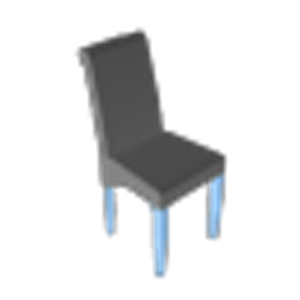}\\
		\includegraphics[width=\linewidth,trim=40 40 40 40,clip]{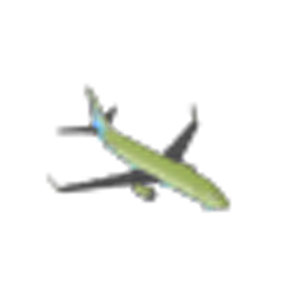}\\
		\caption{Input image}
	\end{subfigure}
	\begin{subfigure}[b]{.74\linewidth}
		\centering
		\includegraphics[width=.32\linewidth,trim=40 40 40 40,clip]{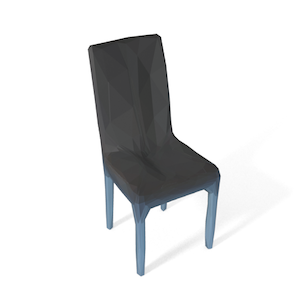}
		\hfill
		\includegraphics[width=.32\linewidth,trim=40 40 40 40,clip]{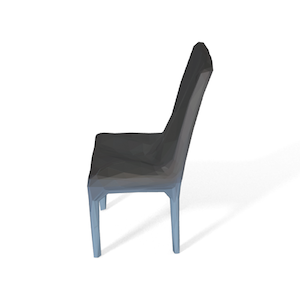}
		\hfill
		\includegraphics[width=.32\linewidth,trim=40 40 40 40,clip]{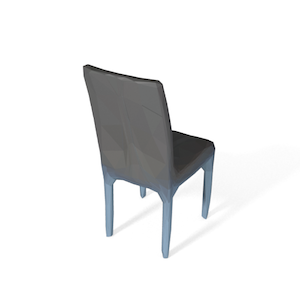}\\
		
		\includegraphics[width=.32\linewidth,trim=40 40 40 40,clip]{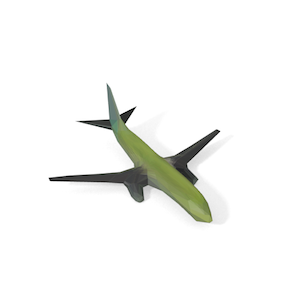}
		\hfill
		\includegraphics[width=.32\linewidth,trim=40 40 40 40,clip]{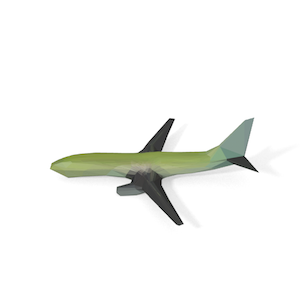}
		\hfill
		\includegraphics[width=.32\linewidth,trim=40 40 40 40,clip]{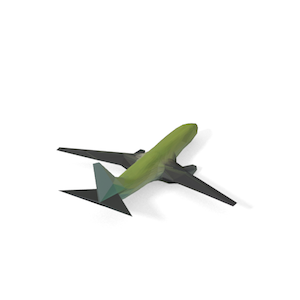}\\
		\caption{Reconstruction results}
	\end{subfigure}
	\vspace{-2mm}
	\caption{Results of colorized mesh reconstruction. }
	\label{fig:color}
	\vspace{-3mm}
\end{figure}

\subsection{Qualitative Results}
\label{sec:quali_results}

\nothing{
Qualitative results that we would like to show:
\begin{itemize}
	\item Pure geometry reconstruction. Comparisons with 
		\begin{itemize}
			\item neural 3D mesh renderer
			\item PTN
			\item Multiview ray consistency 
			\item Pixel2Mesh
		\end{itemize}
	\item Color reconstruction. Comparisons with 
		\begin{itemize}
			\item neural 3D mesh renderer
			\item our naive approach of learning color
		\end{itemize}
	\item Reconstruction using real image (train with more views)
	\item Reconstruction in presence of occlusion (optional)
\end{itemize}	
}


\paragraph{Single-view Mesh Reconstruction.} 

We compare the qualitative results of our approach with that of the state-of-the-art supervised~\cite{wang2018pixel2mesh} and unsupervised~\cite{kato2018neural} mesh reconstruction approaches in Figure~\ref{fig:vis_compare}.
Though Neural 3D Mesh Renderer (N3MR)~\cite{kato2018neural} is able to recover the rough shape, the mesh surface is discontinuous and suffers from a considerable amount of self intersections.
In contrast, our method can faithfully reconstruct fine details of the object, such as the empennage and engines of airplane and the barrel of rifle, while ensuring smoothness of the generated surface.
Though trained in an entirely unsupervised manner, our approach achieves comparable results with the supervised method Pixel2Mesh~\cite{wang2018pixel2mesh}.
For objects that are not genus-0, e.g. the bench in the second row of Figure~\ref{fig:vis_compare}, Pixel2Mesh generates better results than ours (see the armrest). 
The reason is that as our method can only generate genus-0 surface, synthesizing the armrest will lead to large 2D IoU loss.
In contrast, Pixel2Mesh employs 3D Chamfer distance as loss metric which could strongly penalize the missing of armrest in the reconstructed model. 
However, in some cases, our approach can generate even smoother and sharper results than that of \cite{wang2018pixel2mesh}, e.g the lamp base, the engine of airplane and the rifle. Scan-to-mesh distance visualization also shows our results achieve comparable accuracy, in terms of tightness between reconstructed meshes and ground truth.

\paragraph{Color Reconstruction.}

Our method is also capable to recover per-vertex color of the generated mesh based on the input image.
Figure~\ref{fig:color} presents the colorized reconstruction from a single image.
Though the resolution of the input image is rather low ($64 \times 64$), our approach is still able to achieve sharp reconstruction and accurately restore the detailed vertex colors, e.g. the blue color of the airplane tail.

\subsection{Quantitative Evaluations}
\label{sec:quant_results}

\nothing{
\weikai{TODO Comparisons:
\begin{itemize}
	\item Neural 3D mesh renderer
	\item Perspective transformation network
	\item PTN - retrieval based method
	\item Multiview ray consistency 
	\item Pixel2Mesh (optional)
\end{itemize}	
}
}

We show the comparisons on 3D IoU score with different approaches.
Table~\ref{tab:shapenet} lists the statistics for all categories.
As seen in Table~\ref{tab:shapenet}, our mean score surpasses the state-of-the-art N3MR by more than $2.1\%$.
In addition, our approach also achieves the top accuracy for most of the categories.
Our score for the table category is slightly worse than PTN \cite{yan2016perspective} and N3MR.
The main reason is that most of our reconstructed tables contain a 30\textdegree  \ embossment over the main flat surface (see Figure~\ref{fig:vis_120} (b)).
We found all training images are rendered from the viewpoints at 30\textdegree \ on the axis of elevation.
This would lead to an ambiguity of the intended structure as our reconstructed model in fact can generate silhouette that matches the input image well.
We provide an ablation analysis in Section~\ref{sec:ablation} to further validate our model with more comprehensive training views.

\begin{figure}[h]
	\vspace{-10px}
	\begin{subfigure}[b]{.24\linewidth}
		\centering
		\includegraphics[width=\linewidth,trim=40 40 40 40,clip]{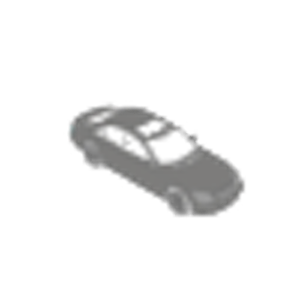}\\
		\vspace{-6px}
		\includegraphics[width=\linewidth,trim=36 36 36 36,clip]{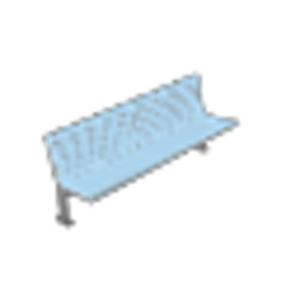}\\
		\vspace{-10px}
		\includegraphics[width=\linewidth,trim=45 45 45 45,clip]{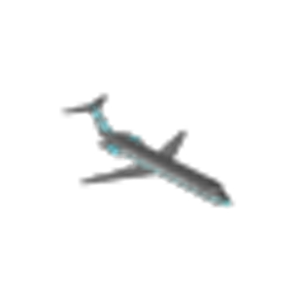}\\
		\caption{Input image}
	\end{subfigure}
	\begin{subfigure}[b]{.24\linewidth}
		\centering
		\includegraphics[width=\linewidth,trim=40 40 40 40,clip]{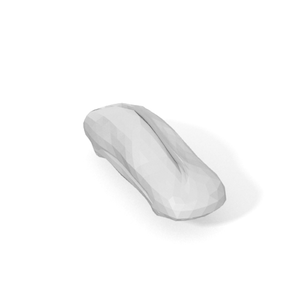}\\
		\vspace{-6px}
		\includegraphics[width=\linewidth,trim=36 36 36 36,clip]{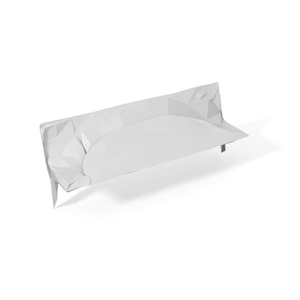}\\
		\vspace{-10px}
		\includegraphics[width=\linewidth,trim=45 45 45 45,clip]{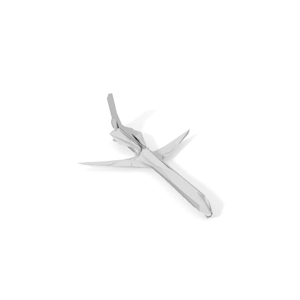}\\
		\caption{w/o $\mathcal{L}_{fl}$}
	\end{subfigure}
	\begin{subfigure}[b]{.24\linewidth}
		\centering
		\includegraphics[width=\linewidth,trim=40 40 40 40,clip]{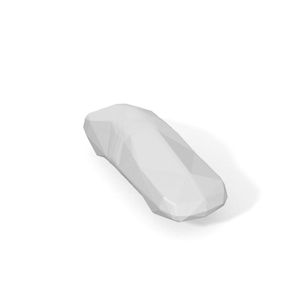}\\
		\vspace{-6px}
		\includegraphics[width=\linewidth,trim=36 36 36 36,clip]{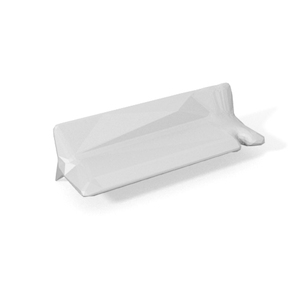}\\
		\vspace{-10px}
		\includegraphics[width=\linewidth,trim=45 45 45 45,clip]{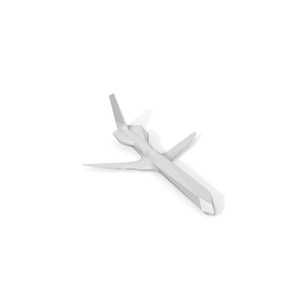}\\
		\caption{w/o $\mathcal{L}_{lap}$}
	\end{subfigure}
	\begin{subfigure}[b]{.24\linewidth}
		\centering
		\includegraphics[width=\linewidth,trim=40 40 40 40,clip]{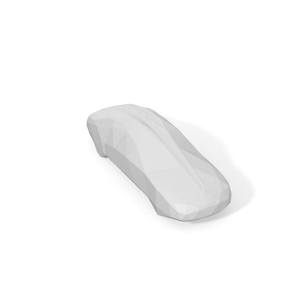}\\
		\vspace{-6px}
		\includegraphics[width=\linewidth,trim=36 36 36 36,clip]{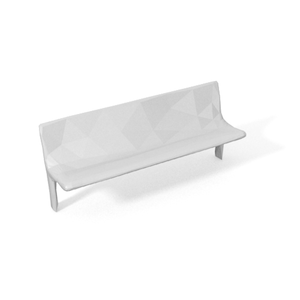}\\
		\vspace{-10px}
		\includegraphics[width=\linewidth,trim=45 45 45 45,clip]{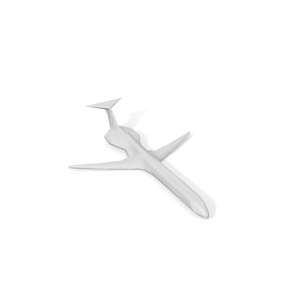}\\
		\caption{Full model}
	\end{subfigure}
	\vspace{-5px}
	\caption{Ablation study of different loss terms.}
	\label{fig:vis_ablation}
\end{figure}
\subsection{Ablation Study}
\label{sec:ablation}

In this section, we conduct controlled experiments to validate the importance of different components.

\paragraph{Loss Terms.}
We first investigate the influence of the two losses, $\loss_{lap}$ and $\loss_{fl}$, that regularize the geometry property of the generated mesh models. 
Figure~\ref{fig:vis_ablation} shows the results of selectively dropping one of the losses.
As seen from the comparisons, removing the Laplacian loss $\loss_{lap}$ would lead to less smooth surface, e.g. the head of airplane, or even discontinuities, e.g. the hole in the bench seat.
Dropping off the flattening loss $\loss_{fl}$ would severely impairs the smoothness of surface and causes intersecting geometry, e.g. the bench back and the tail part of airplane.
However, when the losses are both present, our full model does not suffer from any of the problems, indicating both of the components are necessary for our final performance.  

\begin{figure}[h]
	\begin{subfigure}[b]{.24\linewidth}
		\centering
		\includegraphics[width=\linewidth,trim=40 40 40 40,clip]{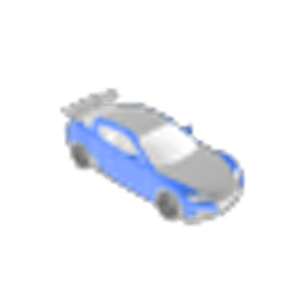}\\
		\vspace{-10px}
		\includegraphics[width=\linewidth,trim=40 40 40 40,clip]{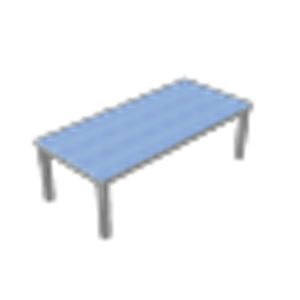}\\
		\caption{Input image}
	\end{subfigure}
	\begin{subfigure}[b]{.24\linewidth}
		\centering
		\includegraphics[width=\linewidth,trim=40 40 40 40,clip]{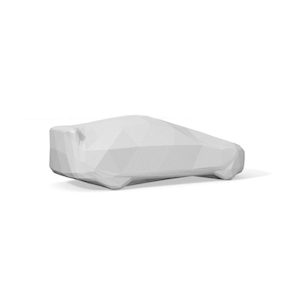}\\
		\vspace{-10px}
		\includegraphics[width=\linewidth,trim=40 40 40 40,clip]{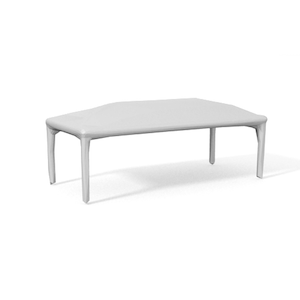}\\
		\label{fig:24view}
		\caption{24 views}
	\end{subfigure}
	\begin{subfigure}[b]{.24\linewidth}
		\centering
		\includegraphics[width=\linewidth,trim=40 40 40 40,clip]{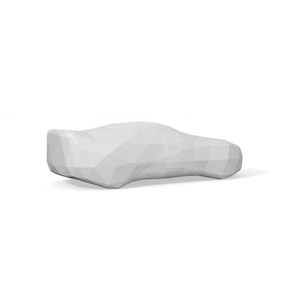}\\
		\vspace{-10px}
		\includegraphics[width=\linewidth,trim=40 40 40 40,clip]{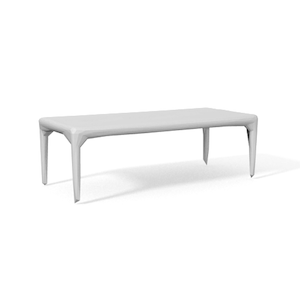}\\
		\label{fig:120view}
		\caption{120 views}
	\end{subfigure}
	\begin{subfigure}[b]{.24\linewidth}
		\centering
		\includegraphics[width=\linewidth,trim=40 40 40 40,clip]{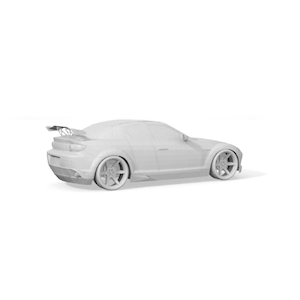}\\
		\vspace{-10px}
		\includegraphics[width=\linewidth,trim=40 40 40 40,clip]{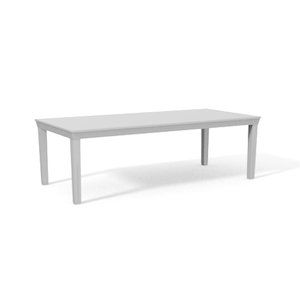}\\
		\caption{{\footnotesize Ground Truth}}
	\end{subfigure}
	\caption{Comparison of results reconstructed from different numbers of training views. With more training views, our approach can achieve reconstruction with significantly higher accuracy.}
	\label{fig:vis_120}
	\vspace{-3mm}
\end{figure}

\paragraph{Training with More Views.}
As discussed in Section~\ref{sec:quant_results}, the existing dataset only contains limited biased views, resulting in a large degree of ambiguity in the reconstructed models. 
To evaluate the full capability of our model, we render the existing models with more comprehensive views. 
In particular, the new training images are rendered from 120 viewpoints, sampled from 5 elevation and 24 azimuth angles.
As shown in Figure~\ref{fig:vis_120}, by training with more views, our model is capable to generate results much closer to the ground truth.
In particular, the embossment of the table model has been completely removed in the new results, indicating the effectiveness of the silhouette supervision obtained from our \modelshort layer.

\begin{figure}[t]
	
	\begin{subfigure}[b]{.24\linewidth}
		\centering
        \includegraphics[width=\linewidth,trim=0 0 0 0,clip]{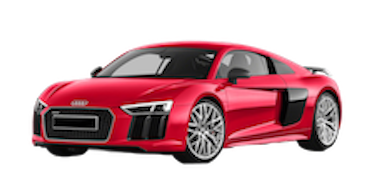}\\
		\vspace{10px}
        \includegraphics[width=\linewidth,trim=20 20 20 20,clip]{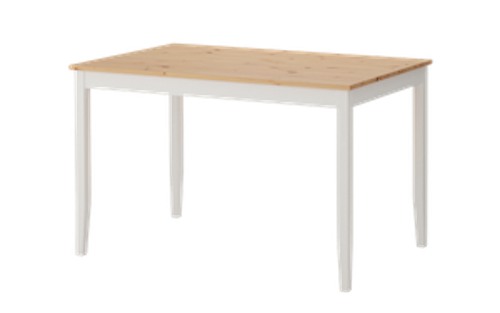}
		\caption{Input image}
	\end{subfigure}
	\begin{subfigure}[b]{.24\linewidth}
		\centering
        \includegraphics[width=\linewidth,trim=0 0 0 0,clip]{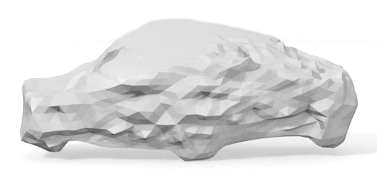}\\
		\vspace{10px}
        \includegraphics[width=\linewidth,trim=20 20 20 20,clip]{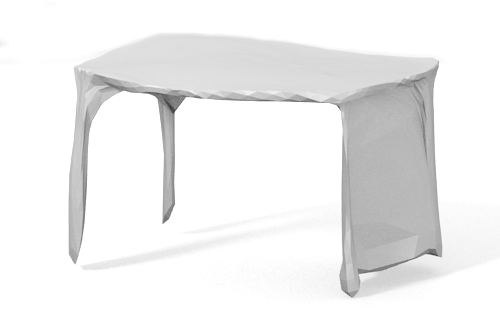}
		\caption{Pixel2Mesh}
	\end{subfigure}
    \begin{subfigure}[b]{.50\linewidth}
		\centering
        \includegraphics[width=.49\linewidth,trim=0 0 0 0,clip]{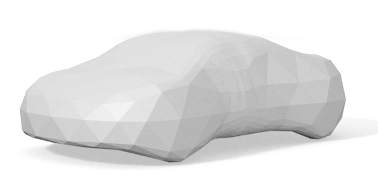}
        \hfill
        \includegraphics[width=.49\linewidth,trim=0 0 0 0,clip]{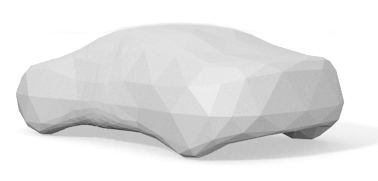} \\
        \vspace{10px}
        \includegraphics[width=.49\linewidth,trim=20 20 20 20,clip]{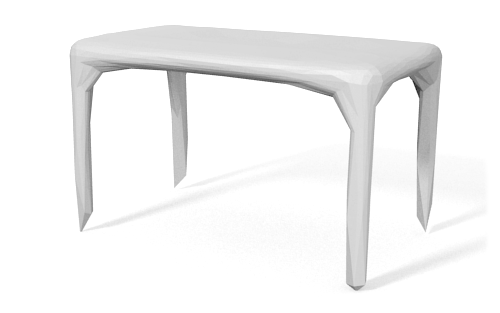}
        \hfill
        \includegraphics[width=.49\linewidth,trim=20 20 20 20,clip]{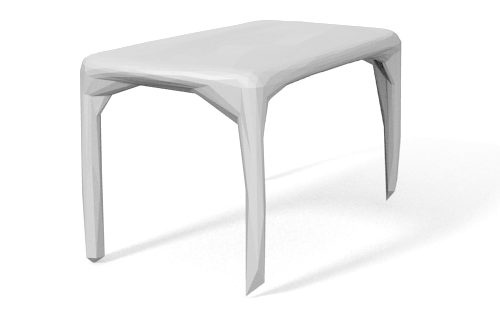} \\
        \caption{SoftRas (ours)}
    \end{subfigure}
	\caption{Comparisons of 3D reconstruction from real images. Our results are shown in two different views.}
	\label{fig:vis_real}
	\vspace{-2mm}
\end{figure}

\paragraph{Reconstruction from Real Image.}
We further evaluate our approach on real images and compare with Pixel2Mesh~\cite{wang2018pixel2mesh}.
Here we employ the model trained on 120 views as proposed in experiment above.
As demonstrated in Figure~\ref{fig:teaser2} (right) and Figure~\ref{fig:vis_real}, though only trained on synthetic data, our model generalizes well to real images in novel views with faithful reconstruction of fine details, e.g. the tail fins of the fighter aircraft.
While for most examples, our unsupervised approach produces comparable results to the supervised method, Pixel2Mesh, we found many cases where our technique can significantly outperform theirs, especially for real world images (see Figure~\ref{fig:vis_real} (b) and (c)).  

\section{Discussion and Future Work}
\label{sec:conclusion}

\nothing{
\begin{wrapfigure}{r}{0.25\textwidth}
	\vspace{-2mm}
	\begin{center}
		\includegraphics[width=0.25\textwidth]{limitPh}
	\end{center}
	\vspace{-5mm}
	\vspace{-1mm}
	\label{fig:fail}
\end{wrapfigure}
}

In this paper, we have presented the first non-parametric differentiable rasterizer (SoftRas) that enables unsupervised learning for high-quality mesh reconstruction from a single image.
We demonstrate that it is possible to properly approximate the forward pass of the discrete rasterization with a differentiable framework.
While many previous works like N3MR~\cite{kato2018neural} seek to provide approximated gradient in the backward propagation but using standard rasterizer in the forward pass, we believe that the consistency between the forward and backward propagations is the key to achieve superior performance.
In addition, we found that proper choice of regularizers plays an important role for producing visually appealing geometry. 
Experiments have shown that our unsupervised approach achieves comparable and in certain cases, even better results to state-of-the-art supervised solutions. 



Limitations and future work. 
As SoftRas only provides silhouette based supervision, ambiguity may arise in our reconstruction when only limited views are available.
Such ambiguity is pronounced when there exists a large planar surface in the object (Figure~\ref{fig:vis_120}(b)).
However, it is possible to resolve this issue by training our model with more comprehensive views (Figure~\ref{fig:vis_120}(c)).    
It would be an interesting future avenue to study the gradient of pixel colors with respect to mesh vertices such that the shading cues can be considered for reconstruction.

\nothing{
\begin{itemize}
	\item Propose a unified framework that could achieve intrinsic color decomposition and leverages both intrinsic color and shading cues to learn the color and further improve the accuracy of geometry.
	\item There may exist other forms of aggregate function that worths exploring.
\end{itemize}
}

\section*{Acknowledgements}
This work was supported in part by the ONR YIP grant N00014-17-S-FO14, the CONIX Research Center, one of six centers in JUMP, a Semiconductor Research Corporation (SRC) program sponsored by DARPA, the Andrew and Erna Viterbi Early Career Chair, the U.S. Army Research Laboratory (ARL) under contract number W911NF-14-D-0005, Adobe, and Sony. The content of the information does not necessarily reflect the position or the policy of the Government, and no official endorsement should be inferred.

{\small
	\bibliographystyle{ieee}
	\bibliography{paper}
}

\ifthenelse{\equal{\final}{0}}
{
}
{}
\end{document}